\documentclass[9pt,twocolumn,twoside]{pnas-altered-arxiv}

\templatetype{pnasresearcharticle} 

\usepackage{graphicx}

\usepackage{textgreek} 
\usepackage{amsmath} 

\usepackage{float}
\usepackage{placeins} 

\usepackage{tablefootnote} 
\usepackage[roman]{parnotes} 
\usepackage{tabulary}
\usepackage{tabularx}

\usepackage{algorithm,algorithmicx}  
\usepackage[noend]{algpseudocode}
\definecolor{mygray}{gray}{0.5}
\definecolor{cblue}{RGB}{8, 85, 153}
\definecolor{darkblue}{RGB}{1, 43, 112}
\definecolor{lblue}{RGB}{204, 255, 255}

\newcommand{\ie}{i.e. }

\newcommand{\ex}[1]{\textbf{$\triangleright$ Ex.\:}}
\newcommand{\define}[1]{\textbf{$\triangleright$ Definition\:}}

\newcommand{\fref}[1]{Fig~\ref{#1}}
\newcommand{\sref}[1]{Sec~\ref{#1}}
\newcommand{\pradi}{PDR \,}

\usepackage{hyperref}

\author[a,1]{W. James Murdoch}
\author[b,1]{Chandan Singh} 
\author[a,2]{Karl Kumbier}
\author[b,c,2]{Reza Abbasi-Asl}
\author[a,b]{Bin Yu}

\affil[a]{UC Berkeley Statistics Dept.}
\affil[b]{UC Berkeley EECS Dept.}
\affil[c]{Allen Institute for Brain Science}

\leadauthor{Murdoch} 

\significancestatement{

The recent surge in interpretability research has led to confusion on numerous fronts. In particular, it is unclear what it means to be \textit{interpretable}, and how to select, evaluate, or even discuss, methods for producing interpretations of machine-learning models. We aim to clarify these concerns by defining interpretable machine learning and constructing a unifying framework for existing methods which highlights the under-appreciated role played by human audiences. 

Within this framework, methods are organized into two classes: model-based and post-hoc. To provide guidance in selecting and evaluating interpretation methods, we introduce three desiderata: predictive accuracy, descriptive accuracy, and relevancy. Using our framework, we review existing work, grounded in real-world studies which exemplify our desiderata, and suggest directions for future work.

}

\authorcontributions{W.M., C.S., K.K., R.A., and B.Y. helped identify important concepts and provided feedback on the paper. W.M and C.S. wrote the paper with contributions from K.K., R.A., and B.Y.}
\authordeclaration{The authors declare no conflict of interest exists.}
\equalauthors{\textsuperscript{1}W.M. and C.S. contributed equally to this work.\\ \\ \textsuperscript{2}K.K. and R.A. contributed equally to this work.}
\correspondingauthor{\textsuperscript{3}To whom correspondence should be addressed: binyu@berkeley.edu}

\keywords{Interpretability $|$ Machine learning $|$ Explainability $|$ Relevancy}

\dates{This manuscript was compiled on \today}
\doi{\url{www.pnas.org/cgi/doi/10.1073/pnas.XXXXXXXXXX}}

\title{Interpretable machine learning: definitions, methods, and applications}

\begin{abstract}
Machine-learning models have demonstrated great success in learning complex patterns that enable them to make predictions about unobserved data. In addition to using models for prediction, the ability to interpret what a model has learned is receiving an increasing amount of attention. However, this increased focus has led to considerable confusion about the notion of interpretability. In particular, it is unclear how the wide array of proposed interpretation methods are related, and what common concepts can be used to evaluate them.
\vspace{3pt} \\
We aim to address these concerns by defining interpretability in the context of machine learning and introducing the \underline{P}redictive, \underline{D}escriptive, \underline{R}elevant (PDR) framework for discussing interpretations. The PDR framework provides three overarching desiderata for evaluation: predictive accuracy, descriptive accuracy and relevancy, with relevancy judged relative to a human audience. Moreover, to help manage the deluge of interpretation methods, we introduce a categorization of existing techniques into model-based and post-hoc categories, with sub-groups including sparsity, modularity and simulatability. To demonstrate how practitioners can use the PDR framework to evaluate and understand interpretations, we provide numerous real-world examples. These examples highlight the often under-appreciated role played by human audiences in discussions of interpretability. Finally, based on our framework, we discuss limitations of existing methods and directions for future work. We hope that this work will provide a common vocabulary that will make it easier for both practitioners and researchers to discuss and choose from the full range of interpretation methods.

\end{abstract}

\begin{document}
\maketitle

\thispagestyle{firststyle}
\abscontentformatted

\section{Introduction}
\label{sec:intro}
\dropcap{M}achine learning (ML) has recently received considerable attention for its ability to accurately predict a wide variety of complex phenomena. However, there is a growing realization that, in addition to predictions, ML models are capable of producing knowledge about domain relationships contained in data, often referred to as interpretations. These interpretations have found uses both in their own right, e.g. medicine \cite{litjens2017survey}, policy-making \cite{brennan2013emergence}, and science \cite{angermueller2016deep,vu2018shared}, as well as in auditing the predictions themselves in response to issues such as regulatory pressure \cite{goodman2016european} and fairness \cite{dwork2012fairness}.

In the absence of a well-formed definition of interpretability, a broad range of methods with a correspondingly broad range of outputs (e.g. visualizations, natural language, mathematical equations) have been labeled as interpretation. This has led to considerable confusion about the notion of interpretability. In particular, it is unclear what it means to interpret something, what common threads exist among disparate methods, and how to select an interpretation method for a particular problem/audience.

In this paper, we attempt to address these concerns. To do so, we first define interpretability in the context of machine learning and place it within a generic data science life cycle. This allows us to distinguish between two main classes of interpretation methods: model-based\footnote{For clarity, throughout the paper we use the term to refer to both machine-learning models and algorithms.} and post hoc. We then introduce the \underline{P}redictive, \underline{D}escriptive, \underline{R}elevant (PDR) framework, consisting of three desiderata for evaluating and constructing interpretations: predictive accuracy, descriptive accuracy, and relevancy, where relevancy is judged by a human audience. Using these terms, we categorize a broad range of existing methods, all grounded in real-world examples\footnote{Examples were selected through a non-exhaustive search of related work.}. In doing so, we provide a common vocabulary for researchers and practitioners to use in evaluating and selecting interpretation methods. We then show how our work enables a clearer discussion of open problems for future research.

\subsection{Defining interpretable machine learning}

On its own, interpretability is a broad, poorly defined concept. Taken to its full generality, to interpret data means to extract information (of some form) from it. The set of methods falling under this umbrella spans everything from designing an initial experiment to visualizing final results. In this overly general form, interpretability is not substantially different from the established concepts of data science and applied statistics.

Instead of general interpretability, we focus on the use of interpretations in the context of ML as part of the larger data-science life cycle. We define interpretable machine learning as the use of machine-learning models for the extraction of \textit{relevant} knowledge about domain relationships contained in data. Here, we view knowledge as being \textit{relevant} if it provides insight for a particular audience into a chosen domain problem. These insights are often used to guide communication, actions, and discovery. Interpretation methods use ML models to produce relevant knowledge about domain relationships contained in data. This knowledge can be produced in formats such as visualizations, natural language or mathematical equations, depending on the context and audience. For instance, a doctor who must diagnose a single patient will want qualitatively different information than an engineer determining if an image classifier is discriminating by race.



\subsection{Background}

Interpretability is a quickly growing field in machine learning, and there have been multiple works examining various aspects of interpretations (sometimes under the heading \textit{explainable AI}). One line of work focuses on providing an overview of different interpretation methods with a strong emphasis on post hoc interpretations of deep learning models \cite{chakraborty2017interpretability, guidotti2018survey}, sometimes pointing out similarities between various methods \cite{lundberg2017unified, ancona2018towards}. Other work has focused on the narrower problem of how interpretations should be evaluated \cite{doshi2017roadmap, gilpin2018explaining} and what properties they should satisfy \cite{lipton2016mythos}. These previous works touch on different subsets of interpretability, but do not address interpretable machine learning as a whole, and give limited guidance on how interpretability can actually be used in data-science life cycles. We aim to do so by providing a framework and vocabulary to fully capture interpretable machine learning, its benefits, and its applications to concrete data problems.


Interpretability also plays a role in other research areas. For example, interpretability is a major topic when considering bias and fairness in ML models \cite{hardt2016equality, boyd2012critical}, with examples given throughout the paper \cite{datta2016algorithmic}. In psychology, the general notions of interpretability and explanations have been studied at a more abstract level \cite{keil2006explanation, lombrozo2006structure}, providing relevant conceptual perspectives. Additionally, we comment on two related areas that are distinct but closely related to interpretability: causal inference and stability.

\paragraph{Causal inference}
Causal inference \cite{imbens2015causal} is a subject from statistics which is related, but distinct, from interpretable machine learning. Causal inference methods focus solely on extracting causal relationships from data, i.e. statements that altering one variable will cause a change in another. In contrast, interpretable ML, and most other statistical techniques, are generally used to describe non-causal relationships, or relationships in observational studies.

In some instances, researchers use both interpretable machine learning and causal inference in a single analysis \cite{basu2018iterative}. One form of this is where the non-causal relationships extracted by interpretable ML are used to suggest potential causal relationships. These relationships can then be further analyzed using causal inference methods, and fully validated through experimental studies. 

\paragraph{Stability}
Stability, as a generalization of robustness in statistics, is a concept that applies throughout the entire data-science life cycle, including interpretable ML). The stability principle requires that each step in the life cycle is stable with respect to appropriate perturbations, such as small changes in the model or data. Recently, stability has been shown to be important in applied statistical problems, for example when trying to make conclusions about a scientific problem \cite{yu2013stability} and in more general settings \cite{hampel2011robust}. Stability can be helpful in evaluating interpretation methods and is a prerequisite for trustworthy interpretations. That is, one should not interpret parts of a model which are not stable to appropriate perturbations to the model and data. This is demonstrated through examples in the text \cite{pimentel2018biclustering, abbasi2018deeptune, basu2018iterative}.

\section{Interpretation in the data science life cycle}
\label{sec:framework}

\begin{figure}[H]
    \centering
    \includegraphics[width=1.0\columnwidth]{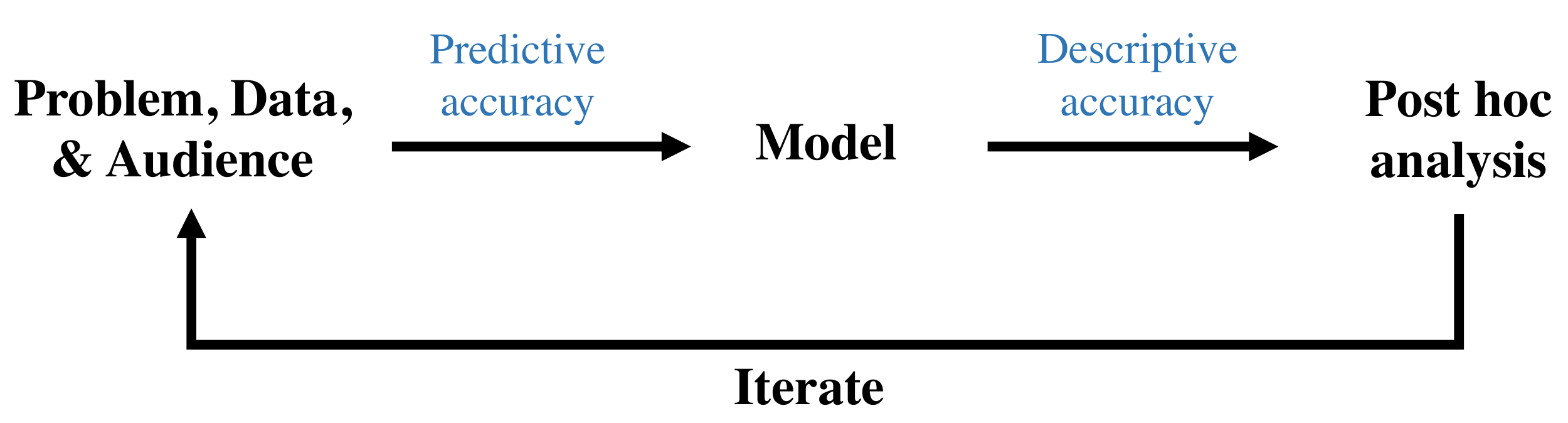}
    \caption{Overview of different stages (black text) in a data-science life cycle where interpretability is important. Main stages are discussed in \sref{sec:framework} and accuracy (blue text) is described in \sref{sec:desiderata}.}
    \label{fig:overview}
\end{figure}

Before discussing interpretation methods, we first place the process of interpretable ML within the broader data-science life cycle. \fref{fig:overview} presents a deliberately general description of this process, intended to capture most data-science problems. What is generally referred to as interpretation largely occurs in the modeling and post hoc analysis stages, with the problem, data and audience providing the context required to choose appropriate methods.

\paragraph{Problem, data, and audience} At the beginning of the cycle, a data-science practitioner defines a domain problem that they would like to understand using data. This problem can take many forms. In a scientific setting, the practitioner may be interested in relationships contained in the data, such as how brain cells in a particular area of the visual system relate to visual stimuli \cite{roe2012toward}. In industrial settings, the problem often concerns the predictive performance or other qualities of a model, such as how to assign credit scores with high accuracy \cite{huang2007credit}, or do so fairly with respect to gender and race \cite{datta2016algorithmic}. The nature of the problem plays a role in interpretability, as the relevant context and audience are essential in determining what methods to use. 

After choosing a domain problem, the practitioner collects data to study it. Aspects of the data-collection process can affect the interpretation pipeline. Notably, biases in the data (\ie mismatches between the collected data and the population of interest) will manifest themselves in the model, restricting one's ability to make interpretations regarding the problem of interest.

\paragraph{Model} Based on the chosen problem and collected data, the practitioner then constructs a predictive model. At this stage, the practitioner processes, cleans, and visualizes data, extracts features, selects a model (or several models) and fits it. Interpretability considerations often come into play in this step related to the choice between simpler, easier to interpret models and more complex, black-box models, which may fit the data better. The model's ability to fit the data is measured through predictive accuracy.

\paragraph{Post hoc analysis} Having fit a model (or models), the practitioner then analyzes it for answers to the original question. The process of analyzing the model often involves using interpretability methods to extract various (stable) forms of information from the model. The extracted information can then be analyzed and displayed using standard data analysis methods, such as scatter plots and histograms. The ability of the interpretations to properly describe what the model has learned is denoted by descriptive accuracy.

\paragraph{Iterate} If sufficient answers are uncovered after the post hoc analysis stage, the practitioner finishes. Otherwise, they update something in the chain (problem, data, and/or model) and iterate, potentially multiple times \cite{box1976science}. Note that they can terminate the loop at any stage, depending on the context of the problem.

\subsection{Interpretation methods within the \pradi framework}
In the framework described above, most interpretation methods fall either in the modeling or post hoc analysis stages. We call interpretability in the modeling stage \textit{model-based interpretability} (\sref{sec:model-based}). This part of interpretability is focused upon constraining the form of ML models so that they readily provide useful information about the uncovered relationships. As a result of these constraints, the space of potential models is smaller, which can result in lower predictive accuracy. Consequently, model-based interpretability is best used when the underlying relationship is relatively simple.

We call interpretability in the post hoc analysis stage \textit{post hoc interpretability} (\sref{sec:post-hoc}). These interpretation methods take a trained model as input, and extract information about what relationships the model has learned. They are most helpful when the underlying relationship is especially complex, and practitioners need to train an intricate, black-box model in order to achieve a reasonable predictive accuracy.

After discussing desiderata for interpretation methods, we investigate these two forms of interpretations in detail and discuss associated methods.

\section{The \pradi desiderata for interpretations}
\label{sec:desiderata}

In general, it is unclear how to select and evaluate interpretation methods for a particular problem and audience. To help guide this process, we introduce the \pradi framework, consisting of three desiderata that should be used to select interpretation methods for a particular problem: predictive accuracy, descriptive accuracy, and relevancy.

\subsection{Accuracy}

The information produced by an interpretation method should be faithful to the underlying process the practitioner is trying to understand. In the context of ML, there are two areas where errors can arise: when approximating the underlying data relationships with a model (predictive accuracy) and when approximating what the model has learned using an interpretation method (descriptive accuracy). For an interpretation to be trustworthy, one should try to maximize both of the accuracies. In cases where the accuracy is not very high, the resulting interpretations may still be useful. However, it is especially important to check their trustworthiness through external validation, such as running an additional experiment.


\subsubsection{Predictive accuracy}
The first source of error occurs during the model stage, when an ML model is constructed. If the model learns a poor approximation of the underlying relationships in the data, any information extracted from the model is unlikely to be accurate. Evaluating the quality of a model's fit has been well-studied in standard supervised ML frameworks, through measures such as test-set accuracy. In the context of interpretation, we describe this error as predictive accuracy.


Note that in problems involving interpretability, one often requires a notion of predictive accuracy that goes beyond just average accuracy. The distribution of predictions matters. For instance, it could be problematic if the prediction error is much higher for a particular class. Moreover, the predictive accuracy should be stable with respect to reasonable data and model perturbations. For instance, one should not trust interpretations from a model which changes dramatically when trained on a slightly smaller subset of the data.

\subsubsection{Descriptive accuracy}
The second source of error occurs during the post hoc analysis stage, when interpretation methods are used to analyze a fitted model. Oftentimes, interpretation methods provide an imperfect representation of the relationships learned by a model. This is especially challenging for complex black-box models such as deep neural networks, which store nonlinear relationships between variables in non-obvious forms.

\define{define} We define \textit{descriptive accuracy}, in the context of interpretation, as the degree to which an interpretation method objectively captures the relationships learned by machine learning models.

\subsubsection{A common conflict: predictive vs descriptive accuracy}
In selecting what model to use, practitioners are often faced with a trade-off between predictive and descriptive accuracy. On the one hand, the simplicity of model-based interpretation methods yields consistently high descriptive accuracy, but can sometimes result in lower predictive accuracy on complex datasets. On the other hand, in complex settings such as image analysis, complicated models generally provide high predictive accuracy, but are harder to analyze, resulting in a lower descriptive accuracy.

\subsection{Relevancy}
When selecting an interpretation method, it is not enough for the method to have high accuracy - the extracted information must also be relevant. For example, in the context of genomics, a patient, doctor, biologist, and statistician may each want different (yet consistent) interpretations from the same model. The context provided by the problem and data stages in \fref{fig:overview} guides what kinds of relationships a practitioner is interested in learning about, and by extension the methods that should be used.

\define{} We define an interpretation to be \textit{relevant} if it provides insight for a particular audience into a chosen domain problem.

Relevancy often plays a key role in determining the trade-off between predictive and descriptive accuracy. Depending on the context of the problem at hand, a practitioner may choose to focus on one over the other. For instance, when interpretability is used to audit a model's predictions, such as to enforce fairness, descriptive accuracy can be more important. In contrast, interpretability can also be used solely as a tool to increase the predictive accuracy of a model, for instance, through improved feature engineering.

\begin{figure}[hbtp!]
    \centering
    \includegraphics[width=0.6\columnwidth]{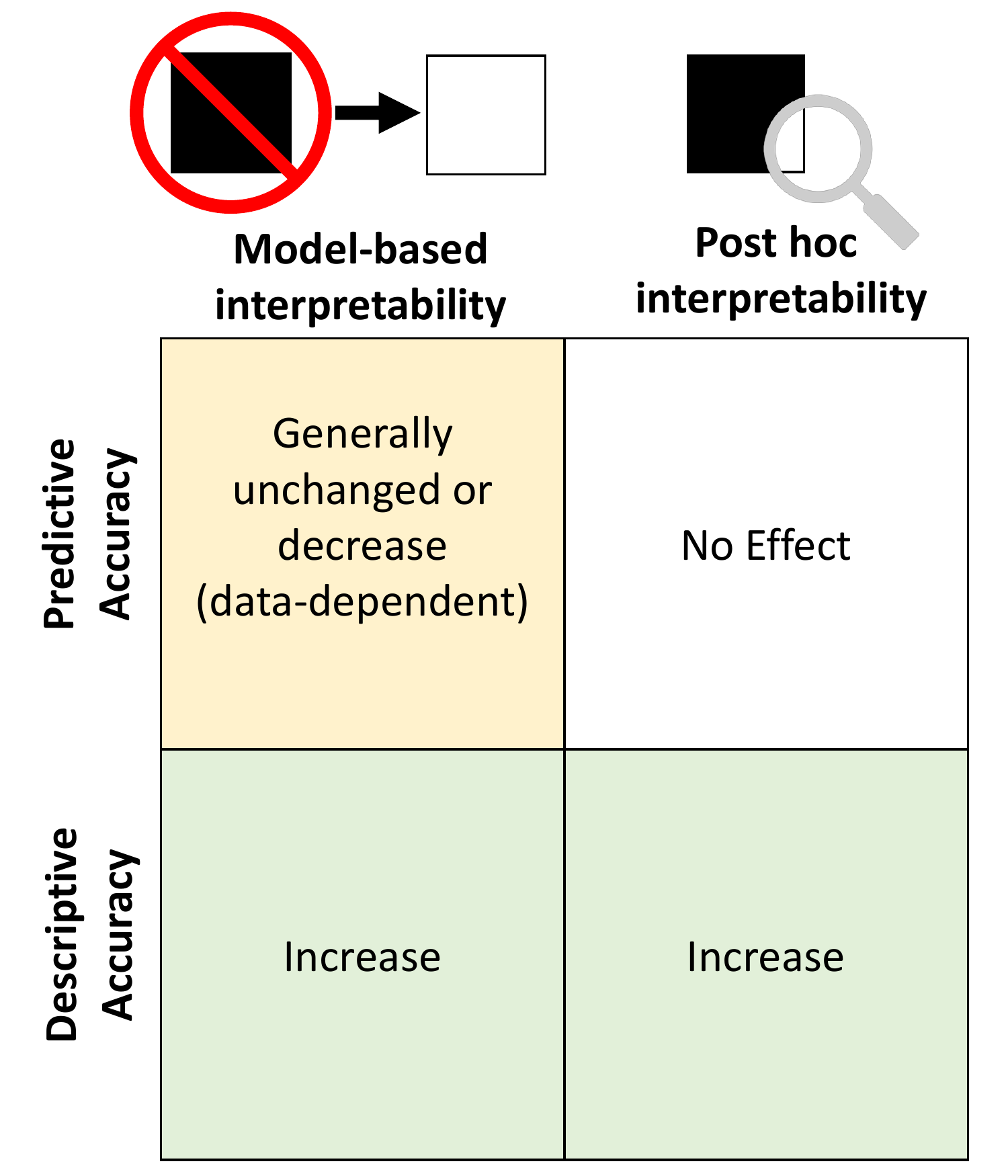} 
    \caption{Impact of interpretability methods on descriptive and predictive accuracies. Model-based interpretability (\sref{sec:model-based}) involves using a simpler model to fit the data which can negatively affect predictive accuracy, but yields higher descriptive accuracy. Post hoc interpretability (\sref{sec:post-hoc}) involves using methods to extract information from a trained model (with no effect on predictive accuracy). These correspond to the model and post hoc stages in \fref{fig:overview}.}
    \label{fig:methods}
\end{figure}

Having outlined the main desiderata for interpretation methods, we now discuss how they link to interpretation in the modeling and post hoc analysis stages in the data-science life cycle. \fref{fig:methods} draws parallels between our desiderata for interpretation techniques introduced in \sref{sec:desiderata} and our categorization of methods in \sref{sec:model-based} and \sref{sec:post-hoc}. In particular, both post hoc and model-based methods aim to increase descriptive accuracy, but only model-based affects the predictive accuracy. Not shown is relevancy, which determines what type of output is helpful for a particular problem and audience.

\section{Model-based interpretability}
\label{sec:model-based}

We now discuss how interpretability considerations come into play in the modeling stage of the data science life cycle (see \fref{fig:overview}). At this stage, the practitioner constructs an ML model from the collected data. We define model-based interpretability as the construction of models that readily provide insight into the relationships they have learned. Different model-based interpretability methods provide different ways of increasing descriptive accuracy by constructing models which are easier to understand, sometimes resulting in lower predictive accuracy. The main challenge of model-based interpretability is to come up with models that are simple enough to be easily understood by the audience, yet sophisticated enough to properly fit the underlying data.

In selecting a model to solve a domain problem, the practitioner must consider the entirety of the \pradi framework. The first desideratum to consider is predictive accuracy. If the constructed model does not accurately represent the underlying problem, any subsequent analysis will be suspect \cite{breiman2001statistical, freedman1991statistical}. Second, the main purpose of model-based interpretation methods is to increase descriptive accuracy. Finally, the relevancy of a model's output must be considered, and is determined by the context of the problem, data, and audience. We now discuss some widely useful types of model-based interpretability methods.

\subsection{Sparsity} When the practitioner believes that the underlying relationship in question is based upon a sparse set of signals, they can impose sparsity on their model by limiting the number of non-zero parameters. In this section, we focus on linear models, but sparsity can be helpful more generally. When the number of non-zero parameters is sufficiently small, a practitioner can interpret the variables corresponding to those parameters as being meaningfully related to the outcome in question, and can also interpret the magnitude and direction of the parameters. However, before one can interpret a sparse parameter set, one should check for stability of the parameters. For example, if the set of sparse parameters changes due to small perturbations in the data set, the coefficients should not be interpreted \cite{lim2016estimation}.

When the practitioner is able to correctly incorporate sparsity into their model, it can improve all three interpretation desiderata. By reducing the number of parameters to analyze, sparse models can be easier to understand, yielding higher descriptive accuracy. Moreover, incorporating prior information in the form sparsity into a sparse problem can help a model achieve higher predictive accuracy and yield more relevant insights. Note that incorporating sparsity can often be quite difficult, as it requires understanding the data-specific structure of the sparsity and how it can be modelled.

Methods for obtaining sparsity often utilize a penalty on a loss function, such as LASSO \cite{tibshirani1996regression} and sparse coding \cite{olshausen1997sparse}, or on a model selection criteria such as AIC or BIC \cite{akaike1987factor, burnham2004multimodel}. Many search-based methods have been developed to find sparse solutions. These methods search through the space of non-zero coefficients using classical subset-selection methods (e.g. orthogonal matching pursuit \cite{pati1993orthogonal}). Model sparsity is often useful for high-dimensional problems, where the goal is to identify key features for further analysis. As a result, sparsity penalties have been incorporated into complex models such as random forests to identify a sparse subset of important features \cite{amaratunga2008enriched}.

In the following example from genomics, sparsity is used to increase the relevancy of the produced interpretations by reducing the number of potential interactions to a manageable level.


\ex{Using sparse canonical correlation analysis in genomics} Identifying interactions among regulatory factors or biomolecules is an important question in genomics. Typical genomic datasets include thousands or even millions of features, many of which are active in specific cellular or developmental contexts. The massive scale of such datasets make interpretation a considerable challenge. Sparsity penalties are frequently used to make the data manageable for statisticians and their collaborating biologists to discuss and identify promising candidates for further experiments. 

For instance, one recent study \cite{pimentel2018biclustering} uses a biclustering approach based on sparse canonical correlation analysis (SCCA) to identify interactions among genomic expression features in \textit{Drosophila melanogaster} (fruit flies) and \textit{Caenorhabditis elegans} (roundworms). Sparsity penalties enable key interactions among features to be summarized in heatmaps which contain few enough variables for a human to analyze. Moreover, this study performs stability analysis on their model, finding it to be robust to different initializations and perturbations to hyperparameters.

\subsection{Simulatability}
A model is said to be simulatable if a human (for whom the interpretation is intended) is able to internally simulate and reason about its entire decision-making process (\ie how a trained model produces an output for an arbitrary input). This is a very strong constraint to place on a model, and can generally only be done when the number of features is low, and the underlying relationship is simple. Decision trees \cite{breiman1984classification} are often cited as a simulatable model, due to their hierarchical decision-making process. Another example is lists of rules \cite{friedman2008predictive, letham2015interpretable}, which can easily be simulated. Due to their simplicity, simulatable models have very high descriptive accuracy. When they can also provide reasonable predictive accuracy, they can be very effective. In the following example, a novel simulatable model is able to produce high predictive accuracy, while maintaining the high levels of descriptive accuracy and relevancy normally attained by rules-based models.


\ex{Rule lists for estimating risk of stroke} In medical practice, when a patient has been diagnosed with atrial fibrillation, caregivers often want to predict the risk that the particular patient will have a stroke in the next year. Moreover, given the potential ramifications of medical decisions, it is important that these predictions are not only accurate, but interpretable to both the caregivers and patients. 

\begin{figure}[H]
    \centering
    \includegraphics[width=\columnwidth]{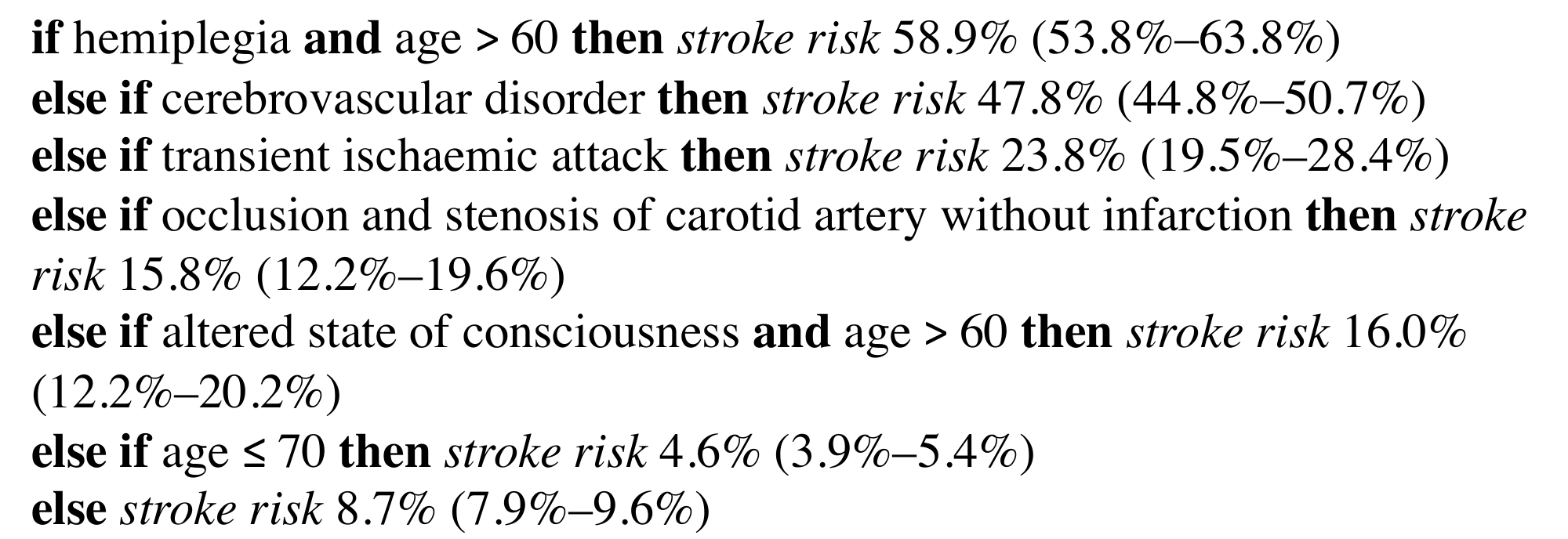}
    \caption{Rule list for classifying stroke risk from patient data (replicated Fig 5 from \cite{letham2015interpretable}). One can easily simulate and understand the relationships between different variables such as age on \textit{stroke risk}. Reprinted with permission from the authors.}
    \label{fig:rule_lists}
\end{figure}

To make the prediction, \cite{letham2015interpretable} uses data from 12,586 patients detailing their age, gender, history of drugs and conditions preceding their diagnosis, and whether they had a stroke within a year of diagnosis. In order to construct a model that has high predictive and descriptive accuracy, \cite{letham2015interpretable} introduce a method for learning lists of if-then rules that are predictive of one year stroke risk. The resulting classifier, displayed in \fref{fig:rule_lists}, requires only seven if-then statements to achieve competitive accuracy, and is easy for even non-technical practitioners to quickly understand.

\subsection{Modularity} We define an ML model to be modular if a meaningful portion(s) of its prediction-making process can be interpreted independently. While modular models are not as easy to understand as sparse or simulatable models, they can still be useful in increasing descriptive accuracy to provide insights into the relationships the model has learned.

A wide array of models satisfy modularity to different degrees. Generalized additive models \cite{hastie1986generalized} force the relationship between variables in the model to be additive. In deep learning, specific methods such as attention \cite{kim2017interpretable} and modular network architectures \cite{andreas2016neural} provide limited insight into a network's inner workings. Probabilistic models can enforce modularity by specifying a conditional independence structure which makes it easier to reason about different parts of a model independently \cite{koller2009probabilistic}.

The following example uses modularity to produce relevant interpretations for use in diagnosing biases in training data.


\ex{Estimating pneumonia risk from patient data} When prioritizing patient care for pneumonia patients in a hospital, one possible method is to predict the likelihood of death within 60 days, and focus on the patients with a higher mortality risk. Given the potential life and death consequences, being able to explain the reasons for hospitalizing a patient or not is very important.

A recent study \cite{caruana2015intelligible} uses a dataset of 14,199 pneumonia patients, with 46 features including from demographics (e.g. age and gender), simple physical measurements (e.g. heart rate, blood pressure) and lab tests (e.g. white blood cell count, blood urea nitrogen). To predict mortality risk, they use a generalized additive model with pairwise interactions, displayed below. The univariate and pairwise terms ($f_j(x_j)$ and $f_{ij}(x_i, x_j)$) can be individually interpreted in the form of curves and heatmaps respectively.
\begin{align}
g(\mathbb{E}[y]) = \beta_0 + \sum_j f_j(x_j) + \sum_{i \neq j} f_{ij}(x_i, x_j)
\end{align}

By inspecting the individual modules, the researchers found a number of counterintuitive properties of their model. For instance, the fitted model learned that having asthma is associated with a lower risk of dying from pneumonia. In reality, the opposite is true - patients with asthma are known to have a higher risk of death from pneumonia. Because of this, in the collected data all patients with asthma received aggressive care, which was fortunately effective at reducing their risk of mortality relative to the general population. 

In this instance, if the model were used without having been interpreted, pneumonia patients with asthma would have have been de-prioritized for hospitalization. Consequently, the use of ML would increase their likelihood of dying. Fortunately, the use of an interpretable model enabled the researchers to identify and correct errors like this one, better ensuring that the model could be trusted in the real world.

\subsection{Domain-based feature engineering} While the type of model is important in producing a useful interpretation, so are the features that are used as inputs to the model. Having more informative features makes the relationship that needs to be learned by the model simpler, allowing one to use other model-based interpretability methods. Moreover, when the features have more meaning to a particular audience, they become easier to interpret.  

In many individual domains, expert knowledge can be useful in constructing feature sets that are useful for building predictive models. The particular algorithms used to extract features are generally domain-specific, relying both on the practitioner's existing domain expertise and insights drawn from the data through exploratory data analysis. For example, in natural language processing, documents are embedded into vectors using tf-idf \cite{ramos2003using} and in computer vision mathematical transformations have been developed to produce useful representations of images \cite{dalal2005histograms}. In the example below, domain knowledge about cloud coverage is exploited to design three simple features that increase the predictive accuracy of a model while maintaining the high descriptive accuracy of a simple predictive model.

\ex{Detecting clouds from satellite imagery} When modelling global climate patterns, an important quantity is the amount and location of arctic cloud coverage. Due to the complex, layered nature of climate models, it is beneficial to have simple, easily auditable, cloud coverage models for use by down-stream climate scientists. 

In \cite{shi2008daytime}, the authors use an unlabeled dataset of arctic satellite imagery to build a model predicting whether each pixel in an image contains clouds or not. Given the qualitative similarity between ice and clouds, this is a challenging prediction problem. By conducting exploratory data analysis and utilizing domain knowledge through interactions with climate scientists, the authors identify three simple features that are sufficient to cluster whether or not images contain clouds. Using these three features as input to quadratic discriminant analysis, they achieve both high predictive accuracy and transparency when compared with expert labels (which were not used in developing the features and the QDA clustering method).

\subsection{Model-based feature engineering} 

There are a variety of automatic approaches for constructing interpretable features. Two examples are unsupervised learning and dimensionality reduction. Unsupervised methods, such as clustering, matrix factorization, and dictionary learning, aim to process unlabelled data and output a description of their structure. These structures often shed insight into relationships contained within the data and can be useful in building predictive models. Dimensionality reduction focuses on finding a representation of the data which is lower-dimensional than the original data. Methods such as principal components analysis \cite{jolliffe1986principal}, independent components analysis \cite{bell1995information}, and canonical correlation analysis \cite{hotelling1936relations} can often identify a few interpretable dimensions, which can then be used as input to a model or to provide insights in their own right. Using fewer inputs can not only improve descriptive accuracy, but can increase predictive accuracy by reducing the number of parameters to fit. In the following example, unsupervised learning (non-negative matrix factorization) is used to represent images in a low-dimensional, genetically meaningful, space.

\ex{Using NMF to find spatial gene expression patterns} Heterogeneity is an important consideration in genomic problems and associated data. In many cases, regulatory factors or biomolecules can play a specific role in one context, such as a particular cell type or developmental stage, and have a very different role in other contexts. Thus, it is important to understand the ``local'' behavior of regulatory factors or biomolecules.

A recent study \cite{wu2016stability}, uses unsupervised learning to learn spatial patterns of gene expression in \textit{Drosophila} (fruit fly) embryos. In particular, they use stability driven nonnegative matrix factorization to decompose images of complex spatial gene expression patterns into a library of $21$ ``principal patterns'', which can be viewed as pre-organ regions This decomposition, which is interpretable to biologists, allows the study of gene-gene interactions in pre-organ regions of the developing embryo.

\section{Post hoc interpretability}
\label{sec:post-hoc}

We now discuss how interpretability considerations come into play in the post hoc analysis stage of the data-science life cycle. At this stage, the practitioner analyzes a trained model in order to provide insights into the learned relationships. This is particularly challenging when the model's parameters do not clearly show what relationships the model has learned. To aid in this process, a variety of post hoc interpretability methods have been developed to provide insight into what a trained model has learned, without changing the underlying model. These methods are particularly important for settings where the collected data is high-dimensional and complex, such as with image data. Once the information has been extracted from the fitted model, it can be analyzed using standard, exploratory data analysis techniques, such as scatter plots and histograms.

When conducting post hoc analysis, the model has already been trained, so its predictive accuracy is fixed. Thus, under the \pradi framework, a researcher must only consider descriptive accuracy and relevancy (relative to a particular audience). Improving on each of these criteria are areas of active research.

Most widely useful post hoc interpretation methods fall into two main categories: prediction-level and dataset-level interpretations, which are sometimes referred to as local and global interpretations, respectively. Prediction-level interpretation methods focus on explaining individual predictions made by models, such as what features and/or interactions led to the particular prediction. Dataset-level approaches focus on the global relationships the model has learned, such as what visual patterns are associated with a predicted response. These two categories have much in common (in fact, dataset-level approaches often yield information at the prediction-level), but we discuss them separately, as methods at different levels are meaningfully different.

\subsection{Dataset-level interpretation}
When a practitioner is interested in more general relationships learned by a model, e.g. relationships that are relevant for a particular class of responses or subpopulation, they use dataset-level interpretations.

\subsubsection{Interaction and feature importances} Feature importance scores, at the dataset-level, try to capture how much individual features contribute, across a dataset, to a prediction. These scores can provide insights into what features the model has identified as important for which outcomes, and their relative importance. Methods have been developed to score individual features in many models including neural networks \cite{olden2004accurate}, random forests, \cite{breiman2001random, strobl2008conditional}, and generic classifiers \cite{altmann2010permutation}.

In addition to feature importances, methods have been developed to extract important interactions between features. Interactions are important as ML models are often highly nonlinear and learn complex interactions between features. Methods exist to extract interactions from a variety of ML models including random forests \cite{basu2018iterative, kumbier2018refining} and neural networks \cite{tsang2017detecting, abbasi2017structural}. In the following example, the descriptive accuracy of random forests is increased by extracting Boolean interactions (a problem-relevant form of interpretation) from a trained model.

\ex{Discovering high-order interactions with iterative random forests (iRFs)} High-order interactions among regulatory factors or genes play an important role in defining cell-type specific behavior in biological systems. As a result, extracting such interactions from genomic data is an important problem in biology. 

A previous line of work considers the problem of searching for biological interactions associated with important biological processes \cite{basu2018iterative, kumbier2018refining}. To identify candidate biological interactions, the authors train a series of iteratively re-weighted RFs and search for stable combinations of features that frequently co-occur along the predictive RF decision paths. This approach takes a step beyond evaluating the importance of individual features in an RF, providing a more complete description of how features influence predicted responses. By interpreting the interactions used in RFs, the researchers identified gene-gene interactions with 80\% accuracy in the \textit{Drosophila} embryo and identify candidate targets for higher-order interactions. 

\subsubsection{Statistical feature importances}
In some instances, in addition to the raw value, we can compute statistical measures of confidence as feature importance scores, a standard technique taught in introductory statistics classes. By making assumptions about the underlying data generating process, models like linear and logistic regression can compute confidence intervals and hypothesis tests for the values, and linear combinations, of their coefficients. These statistics can be helpful in determining the degree to which the observed coefficients are statistically significant. It is important to note that the assumptions of the underlying probabilistic model must be fully verified before using this form of interpretation. Below we present a cautionary example where different assumptions lead to opposing conclusions being drawn from the same dataset.

\ex{Linear models in the Harvard discrimination lawsuit} Here, we consider the lawsuit \textit{Students for Fair Admissions, Inc. v. Harvard} regarding the use of race in undergraduate admissions to Harvard University. Initial reports by Harvard's Office of Institutional Research used logistic regression to model the probability of admission using different features of an applicant's profile, including their race \cite{harvardreportorig}. This analysis found that the coefficient associated with being Asian (and not low income) had a coefficient of -0.418 with a significant p-value (<0.001). This negative coefficient suggested that being Asian had a significant negative association with admission probability.

Subsequent analysis from both sides in the lawsuit attempted to analyze the modeling and assumptions to decide on the significance of race in the model's decision. The plaintiff's expert report \cite{plaintiffreport} suggested that race was being unfairly used by building on the original report from Harvard's Office of Institutional Research. It also incorporates analysis on more subjective factors such as ``personal ratings'' which seem to hurt Asian students' admission. In contrast, the expert report supporting Harvard University \cite{harvardexpertreport} finds that by accounting for certain other variables, the effect of race on Asian students acceptance is no longer significant. Significances derived from statistical tests in regression or logistic regression models at best establish association, but not causation. Hence the analyses from both sides are flawed. This example demonstrates the practical and misleading consequences of statistical feature importances when used inappropriately.

\subsubsection{Visualizations} 

When dealing with high-dimensional datasets, it can be challenging to quickly understand the complex relationships that a model has learned, making the presentation of the results particularly important. To help deal with this, researchers have developed a number of different visualizations which help to understand what a model has learned. For linear models with regularization, plots of regression coefficient paths show how varying a regularization parameter affects the fitted coefficients. When visualizing convolutional neural networks trained on image data, work has been done on visualizing filters \cite{zeiler2014visualizing,olah2017feature}, maximally activating responses of individual neurons or classes \cite{mordvintsev2015deepdream}, understanding intra-class variation \cite{wei2015understanding}, and grouping different neurons \cite{zhang2017interpreting}. For Long Short Term Memory Networks (LSTMs), researchers have focused on analyzing the state vector, identifying individual dimensions that correspond to meaningful features (e.g. position in line, within quotes) \cite{karpathy2015visualizing}, and building tools to track the model's decision process over the course of a sequence \cite{strobelt2016visual}. 

In the following example, relevant interpretations are produced by using maximal activation images for identifying patterns that drive the response of brain cells.

\ex{Understanding visual brain cells via DeepTune} A recent study visualizes learned information from deep neural networks to understand individual brain cells \cite{abbasi2018deeptune}. In this study, macaque monkeys were shown images while the responses of brain cells in their visual system (area V4) were recorded. Neural networks were trained to predict the responses of brain cells to the images. These neural networks produce accurate fits, but provide little insight into what patterns in the images increase the brain cells response without further analysis. To remedy this, the authors introduce DeepTune, a method which provides a visualization, accessible to neuroscientists and others, of the patterns which activate a brain cell. The main intuition behind the method is to optimize the input of a network to maximize the response of a neural network model (which represent a brain cell). 

The authors go on to analyze the major problem of instability. When post hoc visualizations attempt to answer scientific questions, the visualizations must be stable to reasonable perturbations (e.g. the choice of model); if there are changes in the visualization due to the choice of a model, it is likely not meaningful. The authors address this explicitly by fitting eighteen different models to the data and using a stable optimization over all the models to produce a final consensus DeepTune visualization.

\subsubsection{Analyzing trends and outliers in predictions} When interpreting the performance of an ML model, it can be helpful to look not just at the average accuracy, but also at the distribution of predictions and errors. For example, residual plots can identify heterogeneity in predictions, and suggest particular data points to analyze, such as outliers in the predictions, or examples which had the largest prediction errors. Moreover, these plots be used to analyze trends across the predictions. For instance, in the example below, influence functions are able to efficiently identify mislabelled data points.

\ex{Identifying key data points via influence functions} This kind of analysis can also be used to identify mislabeled training data. A recently introduced method \cite{koh2017understanding} uses the classical statistical concept of influence functions to identify points in the training data which contribute to predictions made by ML models. By searching for training data points which contribute the most amount to individual predictions, they were able to find mislabelled data points without having to look at too much data. Correcting these mislabeled training points subsequently improved the test accuracy.

\subsection{Prediction-level interpretation}

Prediction-level approaches are useful when a practitioner is interested in understanding how individual predictions are made by a model. Note that prediction-level approaches can sometimes be aggregated to yield dataset-level insights.

\subsubsection{Feature importance scores} The most popular approach to prediction-level interpretation has involved assigning importance scores to individual features. Intuitively, a variable with a large positive (negative) score made a highly positive (negative) contribution to a particular prediction. In the deep learning literature, a number of different approaches have been proposed to address this problem \cite{springenberg2014striving, sundararajan2016gradients, selvaraju2016grad, baehrens2010explain, smilkov2017smoothgrad, shrikumar2016not, murdoch2017automatic, dabkowski2017real, ribeiro2016should,zintgraf2017visualizing}, with some methods for other models as well \cite{lundberg2018consistent}. These are often displayed in the form of a heat map highlighting important features. Note that feature importance scores at the prediction-level can offer much more information than feature importance scores at the dataset-level. This is a result of heterogeneity in a nonlinear model: the importance of a feature can vary for different examples as a result of interactions with other features. In the following example, feature importance scores are used to increase the descriptive accuracy of black-box models in order to validate their fairness. 


\ex{Auditing predictors for fairness} When using ML models to predict sensitive outcomes, such as whether a person should receive a loan or a criminal sentence, it is important to verify that the algorithm is not discriminating against people based on protected attributes, such as race or gender. This problem is often described as ensuring ML models are ``fair''. In \cite{datta2016algorithmic}, the authors introduce a variable importance measure designed to isolate the contributions of individual variables, such as gender, among a set of correlated variables. 

\begin{figure}
    \centering
    \includegraphics[width=0.7\columnwidth]{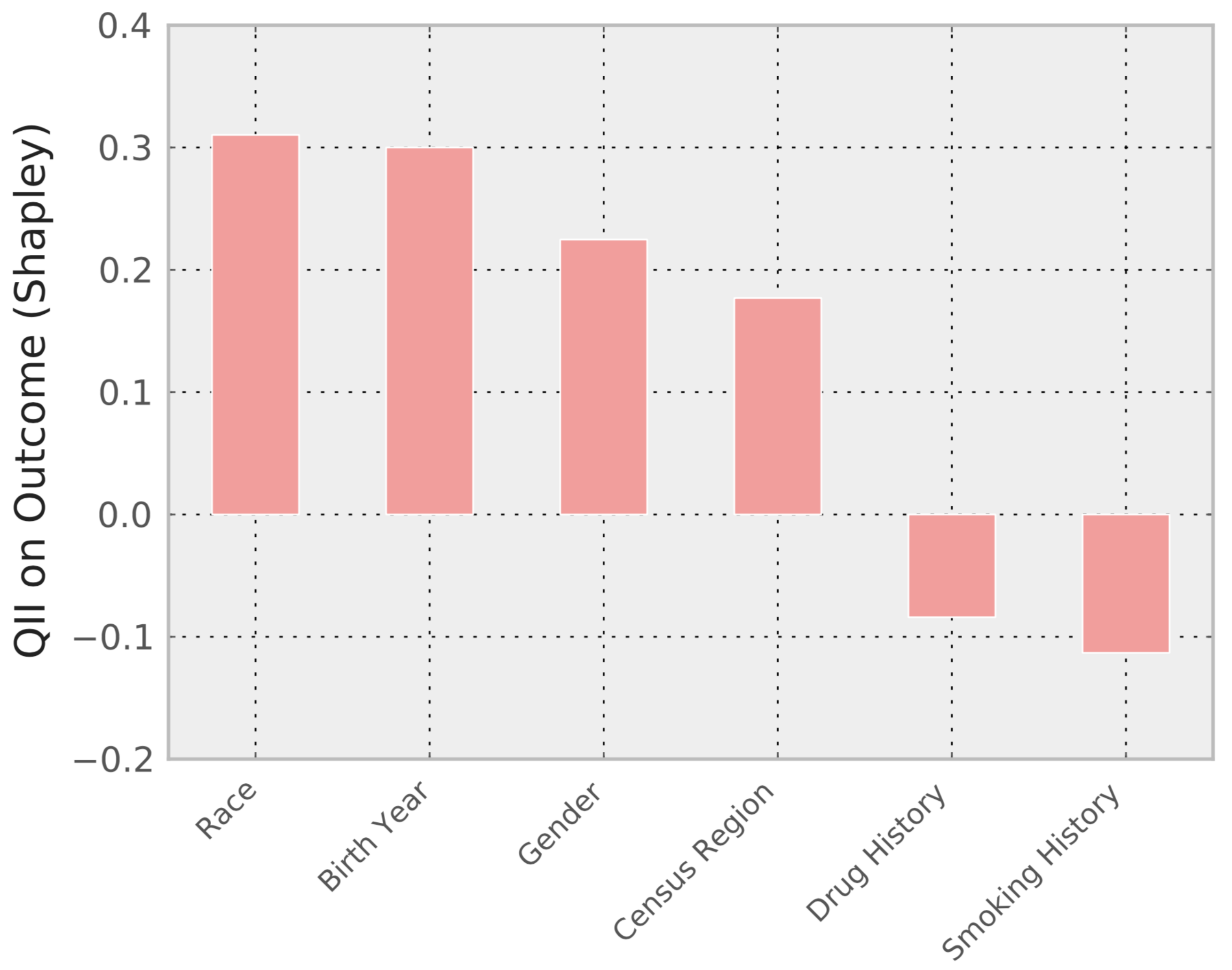}
    \caption{Importance of different predictors in predicting the likelihood of arrest for a particular person. Reprinted with permission from the authors.}
    \label{fig:fairness}
\end{figure}

Based on these variable importance scores, the authors construct transparency reports, such as the one displayed in \fref{fig:fairness}. This figure displays the importance of features used to predict that "Mr. Z" is likely to be arrested in the future (an outcome which is often used in predictive policing), with each bar corresponding to a feature provided to the classifier, and the y axis displaying the importance score for that feature. In this instance, the race feature is the largest value, indicating that the classifier is indeed discriminating based on race. Thus, in this instance, prediction-level feature importance scores are able to identify that a model is unfairly discriminating based on race.

\subsubsection{Alternatives to feature importances} While feature importance scores can provide useful insights, they also have a number of limitations. For instance, they are unable to capture when algorithms learn interactions between variables. There is currently an evolving body of work centered around uncovering and addressing these limitations. These methods focus on explicitly capturing and displaying the interactions learned by a neural network \cite{murdoch2018beyond,singh2018hierarchical}, alternative forms of interpretations such as textual explanations \cite{rohrbach2016grounding}, identifying influential data points \cite{koh2017understanding}, and analyzing nearest neighbors \cite{caruana1999case, papernot2018deep}.

\section{Future work}

Having introduced the \pradi framework for defining and discussing interpretable machine learning, we now leverage it to frame what we feel are the field's most important challenges moving forward. Below, we present open problems tied to each of the paper's three main sections: interpretation desiderata (\sref{sec:desiderata}), model-based interpretability (\sref{sec:model-based}), and post hoc interpretability (\sref{sec:post-hoc}).

\subsection{Measuring interpretation desiderata}

Currently, there is no clear consensus in the community around how to evaluate interpretation methods, although some recent work has begun to address it \cite{doshi2017roadmap, lipton2016mythos, gilpin2018explaining}. As a result, the standard of evaluation varies considerably across different work, making it challenging both for researchers in the field to measure progress, and for prospective users to select suitable methods. Within the \pradi framework, to constitute an improvement, a new interpretation method must improve at least one desideratum (predictive accuracy, descriptive accuracy, or relevancy) without unduly harming the others. While improvements in predictive accuracy are easy to measure, measuring improvements in descriptive accuracy and relevancy remains a challenge. Less important areas for improvement include computational cost and ease of implementation.


\subsubsection{Measuring descriptive accuracy} One way to measure an improvement to an interpretation method is to demonstrate that its output better captures what the ML model has learned, i.e. its descriptive accuracy. However, unlike predictive accuracy, descriptive accuracy is generally very challenging to measure or quantify. As a fall-back, researchers often show individual, cherry-picked, interpretations which seem ``reasonable''. These kinds of evaluations are limited and unfalsifiable. In particular, these results are limited to the few examples shown, and not generally applicable to the entire dataset.

While the community has not settled on a standard evaluation protocol, there are some promising directions. In particular, the use of simulation studies presents a partial solution. In this setting, a researcher defines a simple generative process, generates a large amount of data from that process, and trains their ML model on that data. Assuming a proper simulation setup, a sufficiently powerful model to recover the generative process, and sufficiently large training data, the trained model should achieve near-perfect generalization accuracy. To compute an evaluation metric, they can then check whether their interpretations recover aspects of the original generative process. For example, \cite{tsang2017detecting, tsang2018can} train neural networks on a suite of generative models with certain built-in interactions, and test whether their method successfully recovers them. Here, due to the ML model's near-perfect generalization accuracy, we know that the model is likely to have recovered some aspects of the generative process, thus providing a ground truth against which to evaluate interpretations. In a related approach, when an underlying scientific problem has been previously studied, prior experimental findings can serve as a partial ground truth to retrospectively validate interpretations \cite{basu2018iterative}. 



\subsubsection{Demonstrating relevancy to real-world problems} Another angle for developing improved interpretation methods is to improve the relevancy of interpretations for some audience or problem. This is normally done by introducing a novel form of output, such as feature heatmaps \cite{sundararajan2016gradients}, rationales \cite{lei2016rationalizing}, feature hierarchies \cite{singh2018hierarchical} or identifying important elements in the training set \cite{koh2017understanding}. A common pitfall in the current literature is to focus exclusively on the novel output, ignoring what real-world problems it can actually solve. Given the abundance of possible interpretations, it is particularly easy for researchers to propose novel methods which do not actually solve any real-world problems.

There have been two dominant approaches for demonstrating improved relevancy. The first, and strongest, is to directly use the introduced method in solving a domain problem. For instance, in one example discussed above \cite{basu2018iterative}, the authors evaluated a new interpretation method (iterative random forests) by demonstrating that it could be used to identify meaningful biological Boolean interactions for use in experiments. In instances like this, where the interpretations are used directly to solve a domain problem, their relevancy is indisputable. A second, less direct, approach is the use of human studies, often through services like Amazon's Mechanical Turk. Here, humans are asked to perform certain tasks, such as evaluating how much they trust a model's predictions \cite{singh2018hierarchical}. While challenging to properly construct and perform, these studies are vital to demonstrate that new interpretation methods are, in fact, relevant to any potential practitioners. However, one shortcoming of this approach is that it is only possible to use a general audience of AMT crowdsourced workers, rather than a more relevant, domain-specific audience.


\subsection{Model-based}

Now that we have discussed the general problem of evaluating interpretations, we highlight important challenges for the two main sub-fields of interpretable machine learning: model-based and post hoc interpretability. Whenever model-based interpretability can achieve reasonable predictive accuracy and relevancy, by virtue of its high descriptive accuracy it is preferable to fitting a more complex model, and relying upon post hoc interpretability. Thus, the main focus for model-based interpretability is increasing its range of possible use cases by increasing its predictive accuracy through more accurate models and transparent feature engineering. It is worth noting that sometimes a combination of model-based and post hoc interpretations is ideal. 

\subsubsection{Building accurate and interpretable models} In many instances, model-based interpretability methods fail to achieve a reasonable predictive accuracy. In these cases, practitioners are forced to abandon model-based interpretations in search of more accurate models. Thus, an effective way of increasing the potential uses for model-based interpretability is to devise new modeling methods which produce higher predictive accuracy while maintaining their high descriptive accuracy and relevance. Promising examples of this work include the previously discussed examples on estimating pneumonia risk from patient data \cite{caruana2015intelligible} and Bayesian models for generating rule lists to estimate a patient's risk of stroke \cite{letham2015interpretable}. Detailed directions for this work are suggested in \cite{rudin2018please}.

\subsubsection{Tools for feature engineering} When we have more informative and meaningful features, we can use simpler modeling methods to achieve a comparable predictive accuracy. Thus, methods that can produce more useful features broaden the potential uses of model-based interpretations. The first main category of work lies in improved tools for exploratory data analysis. By better enabling researchers to interact with and understand their data, these tools (combined with domain knowledge) provide increased opportunities for them to identify helpful features. Examples include interactive environments \cite{kluyver2016jupyter, perez2007ipython, RStudio}, tools for visualization \cite{barter2015superheat, wickham2016ggplot2, waskom2014seaborn}, and data exploration tools \cite{mckinney2010data, tidyverse}. The second category falls under unsupervised learning, which is often used as a tool for automatically finding relevant structure in data. Improvements in unsupervised techniques such as clustering and matrix factorization could lead to more useful features. 

\subsection{Post hoc}

In contrast to model-based interpretability, much of post hoc interpretability is relatively new, with many foundational concepts still unclear. In particular, we feel that two of the most important questions to be answered are what an interpretation of an ML model should look like, and how post hoc interpretations can be used. One of the most promising potential uses of post hoc interpretations is to increase the predictive accuracy of a model. In related work, it has been pointed out that in high stakes decisions practitioners should be very careful when applying post hoc methods with unknown descriptive accuracy \cite{rudin2018please}.


\subsubsection{What should an interpretation of a black-box look like} Given a black-box predictor and real-world problem, it is generally unclear what format, or combination of formats, is best to fully capture a model's behavior. Researchers have proposed a variety of interpretation forms, including feature heatmaps \cite{sundararajan2016gradients}, feature hierarchies \cite{singh2018hierarchical} and identifying important elements in the training set \cite{koh2017understanding}. However, in all instances there is a gap between the relatively simple information provided by these interpretations and what the complex model has actually learned. Moreover, it is unclear if any of the current interpretation forms can fully capture a model's behaviour, or if a new format altogether is needed. How to close that gap, while producing outputs relevant to a particular audience/problem, is an open problem.

\subsubsection{Using interpretations to improve predictive accuracy} In some instances, post hoc interpretations uncover that a model has learned relationships a practitioner knows to be incorrect. For instance, prior interpretation work has shown that a binary husky vs. wolf classifier simply learns to identify whether there is snow in the image, ignoring the animals themselves \cite{ribeiro2016should}. A natural question to ask is whether it is possible for the practitioner to correct these relationships learned by the model, and consequently increase its predictive accuracy. Given the challenges surrounding simply generating post hoc interpretations, research on their uses has been limited \cite{ross2017right, zaidan2007using}. However, as the field of post hoc interpretations continues to mature, this could be an exciting avenue for researchers to increase the predictive accuracy of their models by exploiting prior knowledge, independently of any other benefits of interpretations.


\acknow{This research was supported in part by grants ARO W911NF1710005, ONR N00014-16-1-2664, NSF DMS-1613002, and NSF IIS 1741340, an NSERC PGS D fellowship, and an Adobe research award. We thank the Center for Science of Information (CSoI), a US NSF Science and Technology Center, under grant agreement CCF-0939370. Reza Abbasi-Asl would like to thank the Allen Institute founder, Paul G. Allen, for his vision, encouragement and support.}

\showacknow{} 

\FloatBarrier
{

\begin{thebibliography}{10}

\bibitem{litjens2017survey}
Litjens G, et~al. (2017) A survey on deep learning in medical image analysis.
\newblock {\em Medical image analysis} 42:60--88.

\bibitem{brennan2013emergence}
Brennan T, Oliver WL (2013) The emergence of machine learning techniques in
  criminology.
\newblock {\em Criminology \& Public Policy} 12(3):551--562.

\bibitem{angermueller2016deep}
Angermueller C, P{\"a}rnamaa T, Parts L, Stegle O (2016) Deep learning for
  computational biology.
\newblock {\em Molecular systems biology} 12(7):878.

\bibitem{vu2018shared}
Vu MAT, et~al. (2018) A shared vision for machine learning in neuroscience.
\newblock {\em Journal of Neuroscience} pp. 0508--17.

\bibitem{goodman2016european}
Goodman B, Flaxman S (2016) European union regulations on algorithmic
  decision-making and a" right to explanation".
\newblock {\em arXiv preprint arXiv:1606.08813}.

\bibitem{dwork2012fairness}
Dwork C, Hardt M, Pitassi T, Reingold O, Zemel R (2012) Fairness through
  awareness in {\em Proceedings of the 3rd innovations in theoretical computer
  science conference}.
\newblock (ACM), pp. 214--226.

\bibitem{chakraborty2017interpretability}
Chakraborty S, et~al. (2017) Interpretability of deep learning models: a survey
  of results in {\em Interpretability of deep learning models: a survey of
  results}.

\bibitem{guidotti2018survey}
Guidotti R, Monreale A, Turini F, Pedreschi D, Giannotti F (2018) A survey of
  methods for explaining black box models.
\newblock {\em arXiv preprint arXiv:1802.01933}.

\bibitem{lundberg2017unified}
Lundberg SM, Lee SI (2017) A unified approach to interpreting model predictions
  in {\em Advances in Neural Information Processing Systems}.
\newblock pp. 4768--4777.

\bibitem{ancona2018towards}
Ancona M, Ceolini E, Oztireli C, Gross M (2018) Towards better understanding of
  gradient-based attribution methods for deep neural networks in {\em 6th
  International Conference on Learning Representations (ICLR 2018)}.

\bibitem{doshi2017roadmap}
Doshi-Velez F, Kim B (2017) A roadmap for a rigorous science of
  interpretability.
\newblock {\em arXiv preprint arXiv:1702.08608}.

\bibitem{gilpin2018explaining}
Gilpin LH, et~al. (2018) Explaining explanations: An approach to evaluating
  interpretability of machine learning.
\newblock {\em arXiv preprint arXiv:1806.00069}.

\bibitem{lipton2016mythos}
Lipton ZC (2016) The mythos of model interpretability.
\newblock {\em arXiv preprint arXiv:1606.03490}.

\bibitem{hardt2016equality}
Hardt M, Price E, Srebro N, , et~al. (2016) Equality of opportunity in
  supervised learning in {\em Advances in neural information processing
  systems}.
\newblock pp. 3315--3323.

\bibitem{boyd2012critical}
Boyd D, Crawford K (2012) Critical questions for big data: Provocations for a
  cultural, technological, and scholarly phenomenon.
\newblock {\em Information, communication \& society} 15(5):662--679.

\bibitem{datta2016algorithmic}
Datta A, Sen S, Zick Y (2016) Algorithmic transparency via quantitative input
  influence: Theory and experiments with learning systems in {\em Security and
  Privacy (SP), 2016 IEEE Symposium on}.
\newblock (IEEE), pp. 598--617.

\bibitem{keil2006explanation}
Keil FC (2006) Explanation and understanding.
\newblock {\em Annu. Rev. Psychol.} 57:227--254.

\bibitem{lombrozo2006structure}
Lombrozo T (2006) The structure and function of explanations.
\newblock {\em Trends in cognitive sciences} 10(10):464--470.

\bibitem{imbens2015causal}
Imbens GW, Rubin DB (2015) {\em Causal inference in statistics, social, and
  biomedical sciences}.
\newblock (Cambridge University Press).

\bibitem{basu2018iterative}
Basu S, Kumbier K, Brown JB, Yu B (2018) iterative random forests to discover
  predictive and stable high-order interactions.
\newblock {\em Proceedings of the National Academy of Sciences} p. 201711236.

\bibitem{yu2013stability}
Yu B (2013) Stability.
\newblock {\em Bernoulli} 19(4):1484--1500.

\bibitem{hampel2011robust}
Hampel FR, Ronchetti EM, Rousseeuw PJ, Stahel WA (2011) {\em Robust statistics:
  the approach based on influence functions}.
\newblock (John Wiley \& Sons) Vol.{} 196.

\bibitem{pimentel2018biclustering}
Pimentel H, Hu Z, Huang H (2018) Biclustering by sparse canonical correlation
  analysis.
\newblock {\em Quantitative Biology} 6(1):56--67.

\bibitem{abbasi2018deeptune}
Abbasi-Asl R, et~al. (2018) The deeptune framework for modeling and
  characterizing neurons in visual cortex area v4.
\newblock {\em bioRxiv} p. 465534.

\bibitem{roe2012toward}
Roe AW, et~al. (2012) Toward a unified theory of visual area v4.
\newblock {\em Neuron} 74(1):12--29.

\bibitem{huang2007credit}
Huang CL, Chen MC, Wang CJ (2007) Credit scoring with a data mining approach
  based on support vector machines.
\newblock {\em Expert systems with applications} 33(4):847--856.

\bibitem{box1976science}
Box GE (1976) Science and statistics.
\newblock {\em Journal of the American Statistical Association}
  71(356):791--799.

\bibitem{breiman2001statistical}
Breiman L, , et~al. (2001) Statistical modeling: The two cultures (with
  comments and a rejoinder by the author).
\newblock {\em Statistical science} 16(3):199--231.

\bibitem{freedman1991statistical}
Freedman DA (1991) Statistical models and shoe leather.
\newblock {\em Sociological methodology} pp. 291--313.

\bibitem{lim2016estimation}
Lim C, Yu B (2016) Estimation stability with cross-validation (escv).
\newblock {\em Journal of Computational and Graphical Statistics}
  25(2):464--492.

\bibitem{tibshirani1996regression}
Tibshirani R (1996) Regression shrinkage and selection via the lasso.
\newblock {\em Journal of the Royal Statistical Society. Series B
  (Methodological)} pp. 267--288.

\bibitem{olshausen1997sparse}
Olshausen BA, Field DJ (1997) Sparse coding with an overcomplete basis set: A
  strategy employed by v1?
\newblock {\em Vision research} 37(23):3311--3325.

\bibitem{akaike1987factor}
Akaike H (1987) Factor analysis and aic in {\em Selected Papers of Hirotugu
  Akaike}.
\newblock (Springer), pp. 371--386.

\bibitem{burnham2004multimodel}
Burnham KP, Anderson DR (2004) Multimodel inference: understanding aic and bic
  in model selection.
\newblock {\em Sociological methods \& research} 33(2):261--304.

\bibitem{pati1993orthogonal}
Pati YC, Rezaiifar R, Krishnaprasad PS (1993) Orthogonal matching pursuit:
  Recursive function approximation with applications to wavelet decomposition
  in {\em Signals, Systems and Computers, 1993. 1993 Conference Record of The
  Twenty-Seventh Asilomar Conference on}.
\newblock (IEEE), pp. 40--44.

\bibitem{amaratunga2008enriched}
Amaratunga D, Cabrera J, Lee YS (2008) Enriched random forests.
\newblock {\em Bioinformatics} 24(18):2010--2014.

\bibitem{breiman1984classification}
Breiman L, Friedman J, Olshen R, Stone CJ (1984) Classification and regression
  trees.

\bibitem{friedman2008predictive}
Friedman JH, Popescu BE, , et~al. (2008) Predictive learning via rule
  ensembles.
\newblock {\em The Annals of Applied Statistics} 2(3):916--954.

\bibitem{letham2015interpretable}
Letham B, Rudin C, McCormick TH, Madigan D, , et~al. (2015) Interpretable
  classifiers using rules and bayesian analysis: Building a better stroke
  prediction model.
\newblock {\em The Annals of Applied Statistics} 9(3):1350--1371.

\bibitem{hastie1986generalized}
Hastie T, Tibshirani R (1986) Generalized additive models.
\newblock {\em Statistical Science} 1(3):297--318.

\bibitem{kim2017interpretable}
Kim J, Canny JF (2017) Interpretable learning for self-driving cars by
  visualizing causal attention. in {\em ICCV}.
\newblock pp. 2961--2969.

\bibitem{andreas2016neural}
Andreas J, Rohrbach M, Darrell T, Klein D (2016) Neural module networks in {\em
  Proceedings of the IEEE Conference on Computer Vision and Pattern
  Recognition}.
\newblock pp. 39--48.

\bibitem{koller2009probabilistic}
Koller D, Friedman N, Bach F (2009) {\em Probabilistic graphical models:
  principles and techniques}.
\newblock (MIT press).

\bibitem{caruana2015intelligible}
Caruana R, et~al. (2015) Intelligible models for healthcare: Predicting
  pneumonia risk and hospital 30-day readmission in {\em Proceedings of the
  21th ACM SIGKDD International Conference on Knowledge Discovery and Data
  Mining}.
\newblock (ACM), pp. 1721--1730.

\bibitem{ramos2003using}
Ramos J, , et~al. (2003) Using tf-idf to determine word relevance in document
  queries in {\em Proceedings of the first instructional conference on machine
  learning}.
\newblock Vol.{} 242, pp. 133--142.

\bibitem{dalal2005histograms}
Dalal N, Triggs B (2005) Histograms of oriented gradients for human detection
  in {\em Computer Vision and Pattern Recognition, 2005. CVPR 2005. IEEE
  Computer Society Conference on}.
\newblock (IEEE), Vol.{}~1, pp. 886--893.

\bibitem{shi2008daytime}
Shi T, Yu B, Clothiaux EE, Braverman AJ (2008) Daytime arctic cloud detection
  based on multi-angle satellite data with case studies.
\newblock {\em Journal of the American Statistical Association}
  103(482):584--593.

\bibitem{jolliffe1986principal}
Jolliffe I (1986) Principal component analysis.

\bibitem{bell1995information}
Bell AJ, Sejnowski TJ (1995) An information-maximization approach to blind
  separation and blind deconvolution.
\newblock {\em Neural computation} 7(6):1129--1159.

\bibitem{hotelling1936relations}
Hotelling H (1936) Relations between two sets of variates.
\newblock {\em Biometrika} 28(3/4):321--377.

\bibitem{wu2016stability}
Wu S, et~al. (2016) Stability-driven nonnegative matrix factorization to
  interpret spatial gene expression and build local gene networks.
\newblock {\em Proceedings of the National Academy of Sciences}
  113(16):4290--4295.

\bibitem{olden2004accurate}
Olden JD, Joy MK, Death RG (2004) An accurate comparison of methods for
  quantifying variable importance in artificial neural networks using simulated
  data.
\newblock {\em Ecological Modelling} 178(3-4):389--397.

\bibitem{breiman2001random}
Breiman L (2001) Random forests.
\newblock {\em Machine learning} 45(1):5--32.

\bibitem{strobl2008conditional}
Strobl C, Boulesteix AL, Kneib T, Augustin T, Zeileis A (2008) Conditional
  variable importance for random forests.
\newblock {\em BMC bioinformatics} 9(1):307.

\bibitem{altmann2010permutation}
Altmann A, Tolo{\c{s}}i L, Sander O, Lengauer T (2010) Permutation importance:
  a corrected feature importance measure.
\newblock {\em Bioinformatics} 26(10):1340--1347.

\bibitem{kumbier2018refining}
Kumbier K, Basu S, Brown JB, Celniker S, Yu B (2018) Refining interaction
  search through signed iterative random forests.
\newblock {\em arXiv preprint arXiv:1810.07287}.

\bibitem{tsang2017detecting}
Tsang M, Cheng D, Liu Y (2017) Detecting statistical interactions from neural
  network weights.
\newblock {\em arXiv preprint arXiv:1705.04977}.

\bibitem{abbasi2017structural}
Abbasi-Asl R, Yu B (2017) Structural compression of convolutional neural
  networks based on greedy filter pruning.
\newblock {\em arXiv preprint arXiv:1705.07356}.

\bibitem{harvardreportorig}
Office~of Institutional~Research HU (2018) Exhibit 157: Demographics of harvard
  college applicants.
\newblock {\em
  http://samv91khoyt2i553a2t1s05i-wpengine.netdna-ssl.com/wp-content/uploads/2018/06/Doc-421-157-May-30-2013-Report.pdf}
  pp. 8--9.

\bibitem{plaintiffreport}
arcidiacono PS (2018) Exhibit a: Expert report of peter s. arcidiacono.
\newblock {\em
  http://samv91khoyt2i553a2t1s05i-wpengine.netdna-ssl.com/wp-content/uploads/2018/06/Doc-415-1-Arcidiacono-Expert-Report.pdf}.

\bibitem{harvardexpertreport}
Card D (2018) Exhibit 33: Report of david card.
\newblock {\em
  https://projects.iq.harvard.edu/files/diverse-education/files/legal\_-\_card\_report\_revised\_filing.pdf}.

\bibitem{zeiler2014visualizing}
Zeiler MD, Fergus R (2014) Visualizing and understanding convolutional networks
  in {\em European conference on computer vision}.
\newblock (Springer), pp. 818--833.

\bibitem{olah2017feature}
Olah C, Mordvintsev A, Schubert L (2017) Feature visualization.
\newblock {\em Distill} 2(11):e7.

\bibitem{mordvintsev2015deepdream}
Mordvintsev A, Olah C, Tyka M (2015) Deepdream-a code example for visualizing
  neural networks.
\newblock {\em Google Research} 2:5.

\bibitem{wei2015understanding}
Wei D, Zhou B, Torrabla A, Freeman W (2015) Understanding intra-class knowledge
  inside cnn.
\newblock {\em arXiv preprint arXiv:1507.02379}.

\bibitem{zhang2017interpreting}
Zhang Q, Cao R, Shi F, Wu YN, Zhu SC (2017) Interpreting cnn knowledge via an
  explanatory graph.
\newblock {\em arXiv preprint arXiv:1708.01785}.

\bibitem{karpathy2015visualizing}
Karpathy A, Johnson J, Fei-Fei L (2015) Visualizing and understanding recurrent
  networks.
\newblock {\em arXiv preprint arXiv:1506.02078}.

\bibitem{strobelt2016visual}
Strobelt H, et~al. (2016) Visual analysis of hidden state dynamics in recurrent
  neural networks.
\newblock {\em CoRR, abs/1606.07461}.

\bibitem{koh2017understanding}
Koh PW, Liang P (2017) Understanding black-box predictions via influence
  functions.
\newblock {\em arXiv preprint arXiv:1703.04730}.

\bibitem{springenberg2014striving}
Springenberg JT, Dosovitskiy A, Brox T, Riedmiller M (2014) Striving for
  simplicity: The all convolutional net.
\newblock {\em arXiv preprint arXiv:1412.6806}.

\bibitem{sundararajan2016gradients}
Sundararajan M, Taly A, Yan Q (2017) Axiomatic attribution for deep networks.
\newblock {\em ICML}.

\bibitem{selvaraju2016grad}
Selvaraju RR, et~al. (2016) Grad-cam: Visual explanations from deep networks
  via gradient-based localization.
\newblock {\em See https://arxiv. org/abs/1610.02391 v3} 7(8).

\bibitem{baehrens2010explain}
Baehrens D, et~al. (2010) How to explain individual classification decisions.
\newblock {\em Journal of Machine Learning Research} 11(Jun):1803--1831.

\bibitem{smilkov2017smoothgrad}
Smilkov D, Thorat N, Kim B, Vi{\'e}gas F, Wattenberg M (2017) Smoothgrad:
  removing noise by adding noise.
\newblock {\em arXiv preprint arXiv:1706.03825}.

\bibitem{shrikumar2016not}
Shrikumar A, Greenside P, Shcherbina A, Kundaje A (2016) Not just a black box:
  Learning important features through propagating activation differences.
\newblock {\em arXiv preprint arXiv:1605.01713}.

\bibitem{murdoch2017automatic}
Murdoch WJ, Szlam A (2017) Automatic rule extraction from long short term
  memory networks.
\newblock {\em arXiv preprint arXiv:1702.02540}.

\bibitem{dabkowski2017real}
Dabkowski P, Gal Y (2017) Real time image saliency for black box classifiers.
\newblock {\em arXiv preprint arXiv:1705.07857}.

\bibitem{ribeiro2016should}
Ribeiro MT, Singh S, Guestrin C (2016) Why should i trust you?: Explaining the
  predictions of any classifier in {\em Proceedings of the 22nd ACM SIGKDD
  International Conference on Knowledge Discovery and Data Mining}.
\newblock (ACM), pp. 1135--1144.

\bibitem{zintgraf2017visualizing}
Zintgraf LM, Cohen TS, Adel T, Welling M (2017) Visualizing deep neural network
  decisions: Prediction difference analysis.
\newblock {\em arXiv preprint arXiv:1702.04595}.

\bibitem{lundberg2018consistent}
Lundberg SM, Erion GG, Lee SI (2018) Consistent individualized feature
  attribution for tree ensembles.
\newblock {\em arXiv preprint arXiv:1802.03888}.

\bibitem{murdoch2018beyond}
Murdoch WJ, Liu PJ, Yu B (2018) Beyond word importance: Contextual
  decomposition to extract interactions from lstms.
\newblock {\em arXiv preprint arXiv:1801.05453}.

\bibitem{singh2018hierarchical}
Singh C, Murdoch WJ, Yu B (2018) Hierarchical interpretations for neural
  network predictions.
\newblock {\em arXiv preprint arXiv:1806.05337}.

\bibitem{rohrbach2016grounding}
Rohrbach A, Rohrbach M, Hu R, Darrell T, Schiele B (2016) Grounding of textual
  phrases in images by reconstruction in {\em European Conference on Computer
  Vision}.
\newblock (Springer), pp. 817--834.

\bibitem{caruana1999case}
Caruana R, Kangarloo H, Dionisio J, Sinha U, Johnson D (1999) Case-based
  explanation of non-case-based learning methods. in {\em Proceedings of the
  AMIA Symposium}.
\newblock (American Medical Informatics Association), p. 212.

\bibitem{papernot2018deep}
Papernot N, McDaniel P (2018) Deep k-nearest neighbors: Towards confident,
  interpretable and robust deep learning.
\newblock {\em arXiv preprint arXiv:1803.04765}.

\bibitem{tsang2018can}
Tsang M, Sun Y, Ren D, Liu Y (2018) Can i trust you more? model-agnostic
  hierarchical explanations.
\newblock {\em arXiv preprint arXiv:1812.04801}.

\bibitem{lei2016rationalizing}
Lei T, Barzilay R, Jaakkola T (2016) Rationalizing neural predictions.
\newblock {\em arXiv preprint arXiv:1606.04155}.

\bibitem{rudin2018please}
Rudin C (2018) Please stop explaining black box models for high stakes
  decisions.
\newblock {\em arXiv preprint arXiv:1811.10154}.

\bibitem{kluyver2016jupyter}
Kluyver T, et~al. (2016) Jupyter notebooks-a publishing format for reproducible
  computational workflows. in {\em ELPUB}.
\newblock pp. 87--90.

\bibitem{perez2007ipython}
P{\'e}rez F, Granger BE (2007) Ipython: a system for interactive scientific
  computing.
\newblock {\em Computing in Science \& Engineering} 9(3).

\bibitem{RStudio}
{RStudio Team} (2016) {\em RStudio: Integrated Development Environment for R}
  (RStudio, Inc., Boston, MA).

\bibitem{barter2015superheat}
Barter R, Yu B (2015) Superheat: Supervised heatmaps for visualizing complex
  data.
\newblock {\em arXiv preprint arXiv:1512.01524}.

\bibitem{wickham2016ggplot2}
Wickham H (2016) {\em ggplot2: elegant graphics for data analysis}.
\newblock (Springer).

\bibitem{waskom2014seaborn}
Waskom M, et~al. (2014) Seaborn: statistical data visualization.
\newblock {\em URL: https://seaborn. pydata. org/(visited on 2017-05-15)}.

\bibitem{mckinney2010data}
McKinney W, , et~al. (2010) Data structures for statistical computing in python
  in {\em Proceedings of the 9th Python in Science Conference}.
\newblock (Austin, TX), Vol.{} 445, pp. 51--56.

\bibitem{tidyverse}
Wickham H (2017) {\em tidyverse: Easily Install and Load the 'Tidyverse'}.
\newblock R package version 1.2.1.

\bibitem{ross2017right}
Ross AS, Hughes MC, Doshi-Velez F (2017) Right for the right reasons: Training
  differentiable models by constraining their explanations.
\newblock {\em arXiv preprint arXiv:1703.03717}.

\bibitem{zaidan2007using}
Zaidan O, Eisner J, Piatko C (2007) Using “annotator rationales” to improve
  machine learning for text categorization in {\em Human Language Technologies
  2007: The Conference of the North American Chapter of the Association for
  Computational Linguistics; Proceedings of the Main Conference}.
\newblock pp. 260--267.

\end{thebibliography}

}

\end{document}